\documentclass[10pt,twocolumn,letterpaper]{article}

\usepackage{cvpr}              % To produce the CAMERA-READY version
\definecolor{cvprblue}{rgb}{0.21,0.49,0.74}
\usepackage[pagebackref,breaklinks,colorlinks,allcolors=cvprblue]{hyperref}

\usepackage{algorithm2e}
\usepackage{algorithmic}
\usepackage{tabularray}
\usepackage{multirow}  
\usepackage[table]{xcolor}
\definecolor{myrowcolor}{rgb}{0.9,0.9,0.9}
\definecolor{myrowcolor2}{rgb}{0.95,0.95,0.95}
\usepackage{pifont}

\def\paperID{35412} % *** Enter the Paper ID here
\def\confName{CVPR}
\def\confYear{2026}

\title{Towards Generalized Multimodal Homography Estimation}

\author{Jinkun You, Jiaxin Cheng, Jie Zhang, Yicong Zhou\thanks{Corresponding author.}\\
Department of Computer and Information Science, University of Macau\\
Macau, China\\
{\tt\small youjinkun09@gmail.com, yc47434@um.edu.mo, jiezh1997@gmail.com, yicongzhou@um.edu.mo}
}

\begin{document}
\maketitle
\begin{abstract}
Supervised and unsupervised homography estimation methods depend on image pairs tailored to specific modalities to achieve high accuracy. However, their performance deteriorates substantially when applied to unseen modalities. To address this issue, we propose a training data synthesis method that generates unaligned image pairs with ground-truth offsets from a single input image. Our approach renders the image pairs with diverse textures and colors while preserving their structural information. These synthetic data empower the trained model to achieve greater robustness and improved generalization across various domains. Additionally, we design a network to fully leverage cross-scale information and decouple color information from feature representations, thus improving estimation accuracy. Extensive experiments show that our training data synthesis method improves generalization performance. The results also confirm the effectiveness of the proposed network.

% For certain modalities, supervised and unsupervised homography estimation methods require being trained on numerous corresponding image pairs to achieve high accuracy. However, the adaptation to the modalities results in unsatisfactory performance when the trained model handles image pairs with unseen modalities. We propose a training data synthesis method to address this problem. It utilizes the same image to synthesize unaligned image pairs and the ground truth for model training. The image pairs are rendered with various textures and colors at different levels. A a result, the model trained on the synthetic data is robust against modalities and achieves better generalization performance. Furthermore, we design a network to fully utilize the cross-scale information and decouple the color information from the feature, increasing estimation accuracy. Extensive experiments indicate that our training data synthesis method leads to better generalization performance. The results also validate the effectiveness of the proposed network.
\end{abstract}    
\section{Introduction}
\label{sec:intro}

Homography estimation identifies a matrix that characterizes the projective transformation between two images of the same scene captured from different viewpoints. This transformation allows one image to be warped using the estimated matrix, enabling spatial alignment with the other image. Achieving this alignment is crucial for various applications, such as image stitching \cite{nie2023parallax, li2024automatic, cai2025object}, image fusion \cite{zhao2023cddfuse, zhao2024equivariant, wu2025fully}, and guided super-resolution \cite{wang2023learning, ye2023c, you2025hffnet}.

Traditional estimation methods depend on hand-crafted features to match points between image pairs for homography matrix calculation \cite{lowe2004distinctive, rublee2011orb, kim2015dasc}. However, these approaches yield insufficient features in low-texture scenarios. To overcome this limitation, deep learning techniques have been introduced, leveraging their remarkable capabilities for feature extraction and representation across various domains \cite{deng2024enable, you2024two, ou2025mr}. The pioneering deep homography estimation method, inspired by supervised learning, extracts features from image pairs to predict the positional offsets of four corner points \cite{detone2016deep}. These offsets can then be used to compute the homography matrix through the direct linear transform algorithm \cite{dubrofsky2009homography}. During training, the model's weights are adjusted by minimizing the difference between the predicted offsets and the ground truth. To further reduce estimation errors, recent supervised approaches have adopted the inverse compositional Lucas-Kanade framework to refine the predicted offsets iteratively \cite{baker2004lucas, zhao2021deep, cao2022iterative}. These methods estimate offsets multiple times and progressively aggregate the results. Nevertheless, supervised methods require well-aligned image pairs or unaligned pairs with ground-truth offsets for training. The collection of aligned pairs poses a significant challenge, particularly for multimodal images captured by different sensors \cite{li2025mulfs}. Additionally, obtaining ground-truth data in real-world scenarios is often difficult \cite{nie2021unsupervised}. These challenges substantially impede the training of supervised methods.

% Unsupervised methods have received increasing attention recently since they do not need aligned images and ground truth for training. Early approaches optimize the model parameters by maximizing the visual similarity between the image pair 
% \cite{nguyen2018unsupervised,zhang2020content,ye2021motion}. To this end, offsets are predicted to align the image pair before calculating the similarity. Nevertheless, their accuracy is low on multimodal image pairs due to the significant appearance difference. Some approaches utilize image translation techniques to unify the modality and learn the homography estimation simultaneously \cite{arar2020unsupervised, wang2022unsupervised,xu2022rfnet}. They can only handle small deformations, thus achieving poor performance. The self-supervised learning is introduced to tackle this problem. It provides pseudo-unaligned image pairs and the corresponding ground truth, enhancing the homography estimation abilities \cite{zhang2024scpnet, yu2025sshnet}. The image appearance or feature consistency can be enforced to narrow the modality gap, further improving the estimation results \cite{song2024unsupervised, zhang2024scpnet, yu2025sshnet}.

Unsupervised methods have gained significant attention recently because they do not require aligned images or ground truth for training. Early approaches focused on optimizing model parameters by maximizing visual similarity between image pairs \cite{nguyen2018unsupervised, zhang2020content, ye2021motion}. To achieve this, offsets are predicted to align the images prior to calculating their similarity. However, these methods struggle with accuracy on multimodal image pairs due to substantial differences in appearance. Some strategies employ image translation techniques to standardize modalities while simultaneously learning homography estimation \cite{arar2020unsupervised, wang2022unsupervised, xu2022rfnet}. Nonetheless, these methods are limited to handling only small deformations, resulting in low accuracy. Self-supervised learning techniques have been introduced to address this issue. These methods generate pseudo-unaligned image pairs along with the corresponding ground truth, thereby enhancing homography estimation capabilities \cite{zhang2024scpnet, yu2025sshnet}. The modality gap can be narrowed by enforcing consistency in image appearance or feature representation \cite{song2024unsupervised, zhang2024scpnet, yu2025sshnet}. As a result, the estimation results are further improved.

\begin{figure}[t]
    \centering
    \includegraphics[width=1\linewidth]{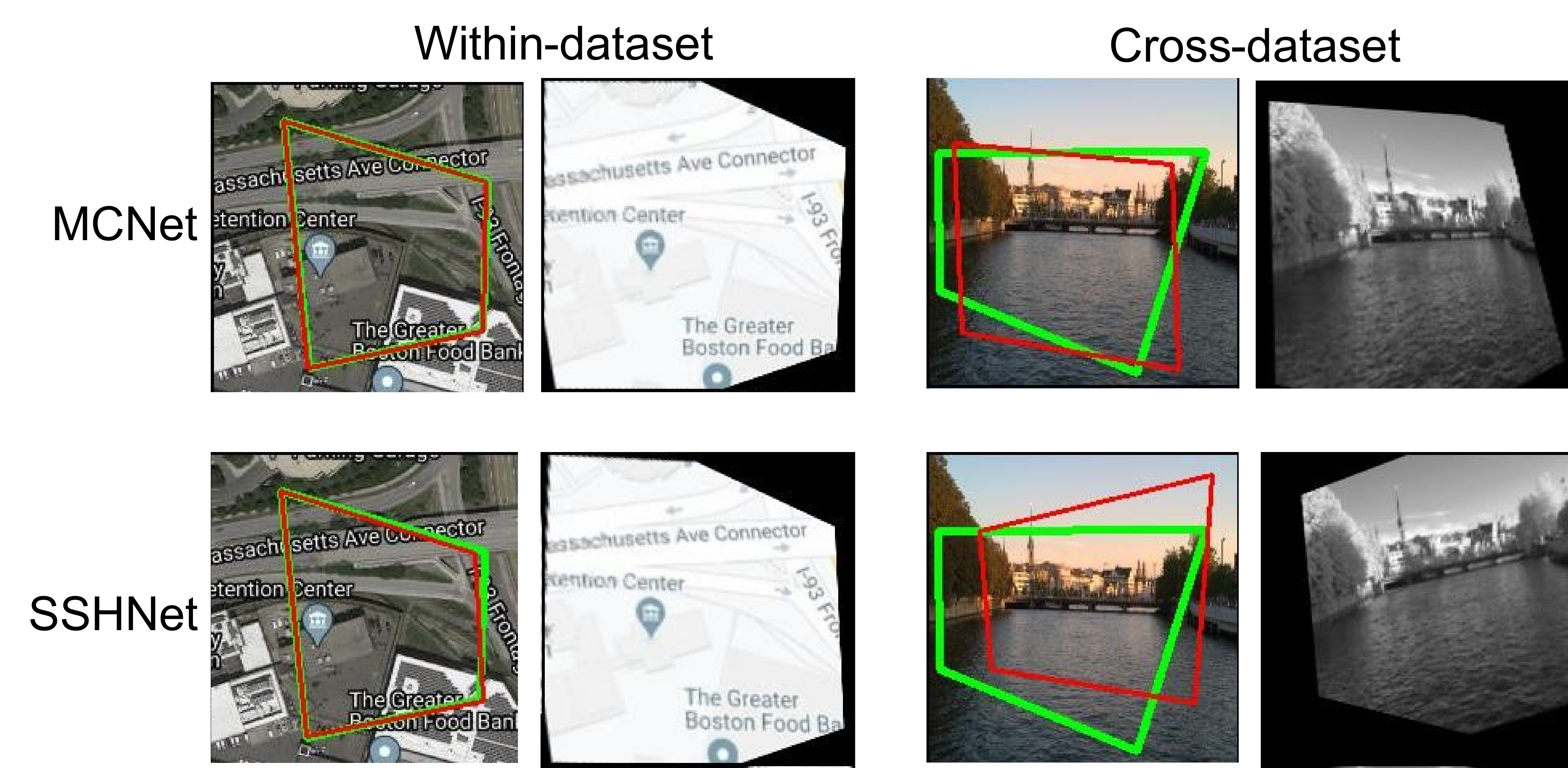}
    \caption{Estimation results. The rows present the results of the supervised MCNet \cite{zhu2024mcnet} and the unsupervised SSHNet \cite{yu2025sshnet}. The first two columns represent within-dataset results, while the remaining columns depict cross-dataset performance. Greater similarity between the green and red quadrilaterals indicates higher accuracy.}
    \label{fig:example_cross_within_acc}
\end{figure}

Supervised and unsupervised homography estimation methods have demonstrated promising accuracy. As illustrated in \cref{fig:example_cross_within_acc}, these models perform admirably when trained and tested on the same dataset. The training data helps adapt the models to the specific modalities involved. However, there are several limitations. First, the trained models exhibit limited generalization capability to other modalities. The estimation accuracy declines when image pairs demonstrate significant appearance differences across modalities. To improve performance, existing approaches require the collection of numerous image pairs from target modalities to learn modality-specific information. This need escalates both time expenditure and labor costs. Second, current methods tend to use features from different scales in isolation. While they effectively harness intra-scale information, they neglect complementary cross-scale information that is beneficial for establishing correspondences between the image pair  \cite{you2025densecrossscaleimagealignment}. Third, the integration of color information within the features can degrade the processing capabilities for multimodal images \cite{pathak2025colors}.

% Supervised and unsupervised homography estimation methods have achieved promising estimation accuracy. As shown in Fig. {\color{red}xxx}, the models have satisfactory estimation performance when being trained and tested on the same dataset. The training data assists in adapting the models to the modalities of the dataset. However, they cannot generalize to other modalities well. The estimation accuracy is decreased since the image pairs differ in appearance across modalities. Existing approaches require collecting numerous image pairs of the target modalities to learn the modality information, thus achieving better performance. This demand increases computational costs and labor intensity. Furthermore, existing methods focus on the intra-scale information and integrate color information into the features. Although multiscale features are beneficial for estimation, existing methods utilize the features of different scales individually. They do not fully utilize the complementary information in the cross-scale features. Images of different modalities have different colors. Therefore, the color information degrades the generalization performance.

To address the above challenges, we introduce a training data synthesis method along with a homography estimation network. Our contributions are as follows:
\begin{itemize}
    \item We propose a training data synthesis method that enables zero-shot multimodal homography estimation. By generating synthetic data with various textures and colors, our approach enhances models' ability to generalize across different modalities. This synthesis method can also be applied to existing datasets, leading to improved generalization performance.
    \item We design a homography estimation network to achieve higher accuracy. It integrates cross-scale information in both top-to-bottom and bottom-to-top directions. Besides, the network decouples color information from feature representations, further enhancing the estimation.
    \item We conduct comprehensive experiments to demonstrate the effectiveness of both the proposed training data synthesis method and the homography estimation network.
\end{itemize}

\section{Related Work}
\label{sec:related}

\textbf{Supervised Homography Estimation.} DeTone \textit{et al.} \cite{detone2016deep} found that parameterizing homography estimation as four-point offsets is more effective. They developed a VGG-style network to regress these offsets based on concatenated image pairs. The model is trained by minimizing the difference between the ground truth and the predicted offsets. Existing supervised methods adopt this training framework. However, regressing offsets just once limits accuracy. To address this, some approaches \cite{erlik2017homography, zhou2019stn, le2020deep} cascade multiple networks for warping and processing input images multiple times. This strategy progressively refines the estimation result and reduces estimation errors, but it also introduces additional parameters. In contrast, the inverse compositional Lucas-Kanade (IC-LK) algorithm maintains model compactness while allowing iterative refinement of the estimation results. Chang \textit{et al.} \cite{chang2017clkn} introduced a differentiable yet untrainable IC-LK layer, enabling effective model training. Building on this, Le \textit{et al.} \cite{zhao2021deep} enhanced performance for multimodal image pairs through brightness-consistent feature extraction. Cao \textit{et al.} \cite{cao2022iterative} proposed a trainable iterative framework to fully leverage the data-driven capabilities of deep learning. They subsequently designed an attention mechanism to capture correspondences across various ranges \cite{cao2023recurrent}. Zhu \textit{et al.} \cite{zhu2024mcnet} integrated the iterative strategy into multiscale correlation searching, resulting in improved estimation accuracy and efficiency.
% DeTone \textit{et al.} \cite{detone2016deep} found that it is more effective to parameterize the homography estimation as four-point offsets. They designed a VGG-style network to regress the offsets from concatenated image pairs.  The difference between the ground truth and the prediction is minimized to train the model. Existing supervised methods follow this training framework. However, regressing offsets only once leads to limited accuracy. Some approaches \cite{erlik2017homography, zhou2019stn, le2020deep} cascade multiple networks to warp the input images or multiscale features several times. This refinement strategy reduces estimation errors progressively but introduces extra parameters. The inverse compositional Lucas-Kanade (IC-LK) algorithm can maintain model size while refining the estimation result iteratively. Chang \textit{et al.} \cite{chang2017clkn} designed a differentiable IC-LK layer to enable model training. Le \textit{et al.} \cite{zhao2021deep} further boosted the performance on multimodal image pairs by brightness-consistent feature extraction. Cao \textit{et al.} \cite{cao2022iterative} proposed a trainable iteration framework to fully leverage the data-driven capabilities of deep learning. Subsequently, they employed the attention mechanism to capture the correspondence of image pairs across various ranges \cite{cao2023recurrent}. Zhu \textit{et al.} \cite{zhu2024mcnet} integrated the iteration strategy into multiscale correlation searching, achieving improved estimation accuracy and efficiency.

\textbf{Unsupervised Homography Estimation.} 
Ground-truth offsets are often unavailable in real-world scenarios. This issue prompts increased interest in unsupervised methods since they do not require ground truth data for model training. Nguyen \textit{et al.} \cite{nguyen2018unsupervised} introduced the first unsupervised method. This method featured a differentiable warping layer and utilized photometric loss to minimize the visual discrepancies between image pairs. To enhance estimation accuracy, Zhang \textit{et al.} \cite{zhang2020content} implemented attention maps to remove outliers and maximized similarity for aligned feature maps. Ye \textit{et al.} \cite{ye2021motion} developed a warp-equivariant feature extractor to reduce the rank of the homography flow. However, these methods are primarily constrained to unimodal images. While some methods have utilized modality transfer to work with multimodal image pairs, they have achieved limited accuracy. To address this challenge, Zhang \textit{et al.} \cite{zhang2024scpnet} proposed a self-supervised framework that generates unaligned image pairs and corresponding ground truth from each training image. These generated data significantly enhance the unimodal estimation capabilities. The features \cite{zhang2024scpnet, song2024unsupervised} or modalities \cite{yu2025sshnet} of the image pair are aligned to allow for multimodal homography estimation.
% Ground-truth offsets are almost unavailable in the real world. Unsupervised methods have received increasing attention since they do not need the ground truth to update the model parameters. Nguyen \textit{et al.} \cite{nguyen2018unsupervised} proposed the first unsupervised method. They designed a differentiable warping layer and used the photometric loss to minimize the visual difference between the image pair. To increase estimation accuracy, Zhang \textit{et al.} \cite{zhang2020content} removed outliers via attention maps and maximized the similarity for aligned feature maps. Ye \textit{et al.} \cite{ye2021motion} devised a Warp-equivariant feature extractor to reduce the rank of the homography flow. The aforementioned methods can only handle RGB images. Some approaches exploited the modality transfer techniques to handle multimodal image pairs but achieved limited accuracy \cite{arar2020unsupervised, wang2022unsupervised, xu2022rfnet}. To tackle this issue, Zhang \textit{et al.} \cite{zhang2024scpnet} introduced a self-supervised framework to generate unaligned image pairs and the ground truth from each training image. These generated training data help improve the unimodal estimation capabilities. The deep features or appearance of the multimodal image pair can be aligned to enable multimodal homogrpahy estimation \cite{zhang2024scpnet,song2024unsupervised}. Yu \textit{et al.} proposed a Transformer network to align the modality of the image pair \cite{yu2025sshnet}.

\textbf{Style Transfer.} 
A style transfer network merges the style of a given template image with the content of another image. Gatys \textit{et al.} \cite{gatys2016image} introduced the first convolutional neural network capable of extracting style features and integrating them with content features. Chen \textit{et al.} \cite{chen2021artistic} designed an internal-external network to tackle issues of color imbalance and pattern repetition. Cheng \textit{et al.} \cite{cheng2021style} developed a loss function to reduce model bias toward specific styles. Zhang \textit{et al.} \cite{zhang2023inversion} enhanced style capture by leveraging text prompts within a pretrained diffusion model. Chung \textit{et al.} \cite{chung2024style} transformed self-attention into cross-attention, eliminating the need to train diffusion models.
% A style transfer network combines the style of a given template image with the content of another image. Gatys \textit{et al.} \cite{gatys2016image} proposed the first convolutional neural network to extract the style feature and integrate it with the content feature. Chen \textit{et al.} \cite{chen2021artistic} designed an internal-external network to address the color imbalance and pattern repetition problems. Cheng \textit{et al.} \cite{cheng2021style} designed a loss function to mitigate the model bias towards specific styles. Zhang \textit{et al.} \cite{zhang2023inversion} leverages the text prompt to enhance the capture of style in the pretrained diffusion model. Chung \textit{et al.} \cite{chung2024style} modified the self-attention to the cross-attention, avoiding the training of diffusion models for style transfer.

\section{Proposed Method}
\label{sec:proposed}

\begin{figure*}[t]
    \centering
    \includegraphics[width=0.98 \linewidth]{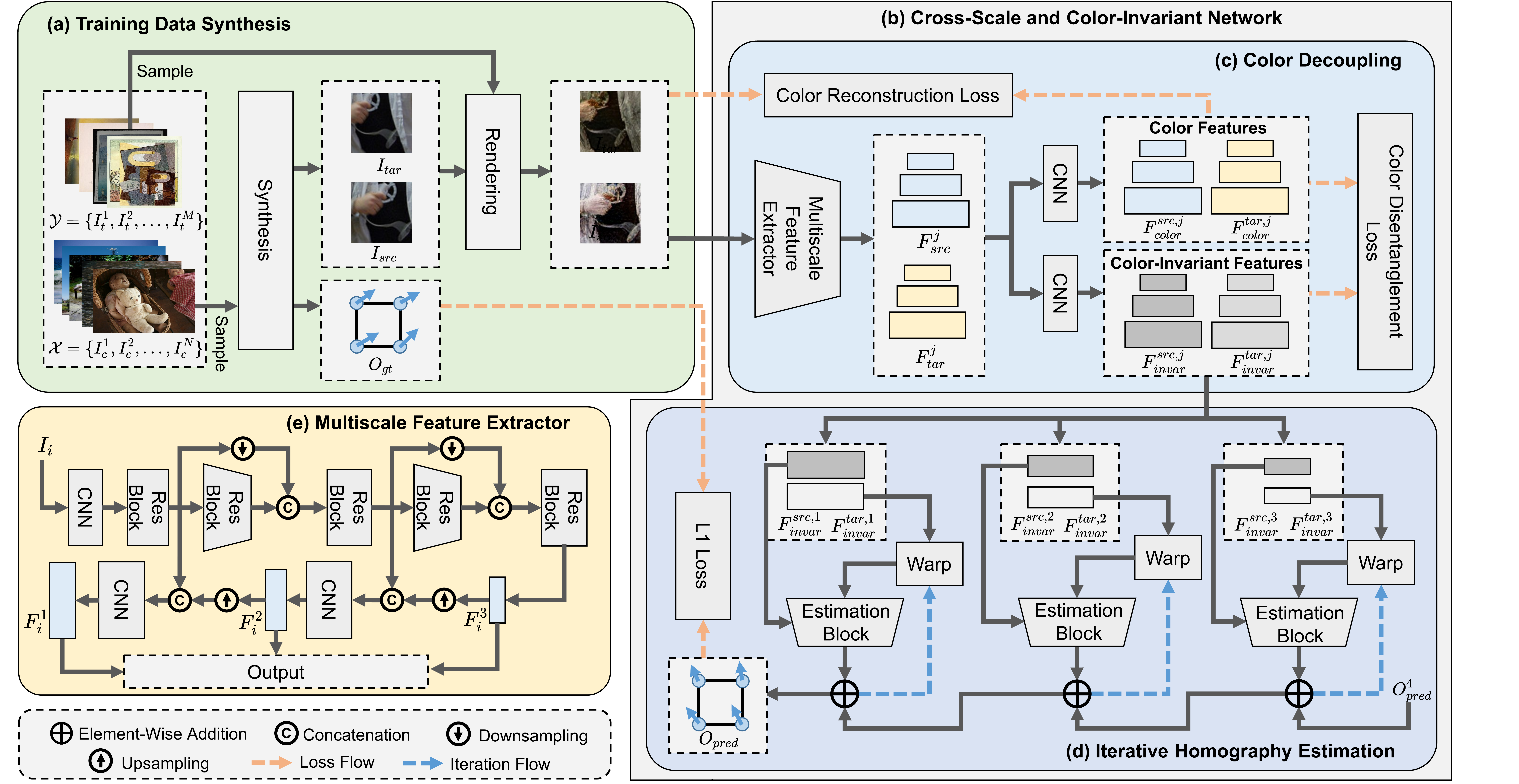}
    \caption{Illustration of the training data synthesis and the homography estimation network. (a) Training Data Synthesis can be applied to a public RGB dataset for zero-shot homography estimation or integrated with existing datasets to enhance generalization. (b) Cross-Scale and Color-Invariant Network (CCNet) integrates cross-scale information into the extracted features while decoupling color information from the feature representations. (c) Color Decoupling. (d) Iterative Homography Estimation. (e) Multiscale Feature Extractor.}
    \label{fig:overall_framework}
\end{figure*}

\subsection{Overall Framework}
\cref{fig:overall_framework}  illustrates the proposed training data synthesis method and the homography estimation network. The synthetic data are utilized for model training to achieve zero-shot estimation and enhance generalization performance. The model training is formulated as:
\begin{equation}
    \theta^*=\max_{\theta}\mathrm{P}(\mathrm{Net}(I_{src},I_{tar}, \theta), O_{gt}),
\end{equation}
where $I_{src}\in\mathbb{R}^{3\times S\times S}$, $I_{tar}\in\mathbb{R}^{3\times S\times S}$, and $O_{gt}\in\mathbb{R}^{4\times 2}$ denote the synthetic source image, target image, and groud-truth offsets, respectively; $\mathrm{Net}(\cdot,\cdot)$ denotes the estimation model and $\theta$ represents its parameters; $\mathrm{P}(\cdot,\cdot)$ evaluates the estimation accuracy. The final objective is given by
\begin{equation}
    \max\mathrm{P}(\mathrm{Net}(I_{src}^\prime,I_{tar}^\prime,\theta^*),O_{gt}^\prime),
\end{equation}
where $I_{src}^\prime$ and $I_{tar}^\prime$ are the image pairs with unseen modalities; $O_{gt}^\prime$ is the ground-truth offsets. The proposed network integrates cross-scale information and decouples color information for higher estimation accuracy. It also employs the iterative strategy to refine the estimation results.

% \RestyleAlgo{ruled}
% \begin{algorithm}
%     \SetKwInOut{Input}{Input}
%     \SetKwInOut{Output}{Output}

%     \Input{content image set $\mathcal{X}=\{I_{c}^1,I_{c}^2,...,I_{c}^N\}$; template image set $\mathcal{Y}=\{I_{t}^1,I_{t}^2,...,I_{t}^M\}$; patch size $S$; margin size $S_m$; maximum perturbation $p$; maximum smoothing weight $\beta$; style transfer network $\mathrm{Net}_s$.}
%     \Output{source image $I_{src}$; target image $I_{tar}$; ground-truth offsets $O_{gt}$.}

%      Randomly sample $I_{c}\in\mathbb{R}^{3\times H\times W}$ from $\mathcal{X}$\;
    
%      Randomly sample $I_{t}^i,I_{t}^j\in\mathbb{R}^{3\times H\times W}$ from $\mathcal{Y}$\;
%      $O_{gt}\gets$torch.randint($-p$, $p+1$, [4,2])\;
     
%      $x\gets$torch.randint($S_m$, $W-S_m-S+1$)\;
%      $y\gets$torch.randint($S_m$, $H-S_m-S+1$)\;
%      $I_{patch}\gets\mathrm{Crop}(I_c,x,y,S_m+S)$\;
    
%      $\alpha_j,\alpha_k\gets$ torch.rand(0,1), torch.rand(0,1)\;
%      $I_{src}\gets \alpha_i\cdot I_{patch}+(1-\alpha_j)\cdot\mathrm{Net}_{s}(I_{patch},I_t^i)$\;
%      $I_{tar}\gets \alpha_j\cdot I_{patch}+(1-\alpha_k)\cdot\mathrm{Net}_{s}(I_{patch},I_t^j)$\;
     
%      $\beta_j,\beta_k\gets$ torch.rand(0,$\beta$), torch.rand(0,$\beta$)\;
%      $I_{src},I_{tar}\gets \mathrm{Smooth}(I_{src},\beta_i), \mathrm{Smooth}(I_{tar},\beta_j)$\;
    
%      $I_{src}\gets\mathrm{Warp}(I_{src},O_{gt})$\;
%      $I_{src}\gets\mathrm{Crop}(I_{src},S_m,S_m,s)$\;
%      $ I_{tar}\gets\mathrm{Crop}(I_{tar},S_m,S_m,s)$\;
%     \caption{Training data synthesis}
%     \label{alg:data_synthesis}
% \end{algorithm}

\subsection{Training Data Synthesis}
As observed in \cref{fig:example_cross_within_acc}, textures and colors vary across different modalities. Therefore, the homography estimation model must be robust to changes in textures and colors. Style transfer networks can effectively blend texture and color information from one image into another, inspiring us to use this approach to render the same image in various styles. These rendered images are then utilized for training data synthesis, providing a diverse array of textures and colors while preserving the structural information.

The synthetic training data consist of the source image $I_{src}\in\mathbb{R}^{3\times S\times S}$, the target image $I_{tar}\in\mathbb{R}^{3\times S\times S}$, and the ground-truth offsets $O_{gt}\in\mathbb{R}^{4\times 2}$. 
%%%%%%%%%%%%%%%%%%%%%%%%%%%%%%%%%%%%%%%% algorithm %%%%%%%%%%%%%%%%%%%%%%%%%%%%%%%%%%%%%%%
% The details of the synthesis process are outlined in \cref{alg:data_synthesis}. 
Initially, a content image $I_c\in\mathbb{R}^{3\times H\times W}$ is randomly sampled from the content dataset $\mathcal{X}=\{I_c^1,I_c^2,...,I_c^N\}$. A patch is then obtained by cropping $I_c$
\begin{equation}
    I_{patch}=\mathrm{Crop}(I_c,x,y,S_m+S),
\end{equation}
where $S_m$ is the margin size; $I_{patch}\in\mathbb{R}^{3\times (S_m+S)\times (S_m+S)}$ represents the cropped patch with its top-left corner positioned at $(x,y)$ in $I_c$; $\textrm{Crop}(\cdot)$ is the cropping function. Next, aligned source and target images are generated by rendering $I_{patch}$ in different styles. Two template images, $I_t^i$ and $I_t^j$, are randomly selected from the template dataset $\mathcal{Y}=\{I_t^1,I_t^2,...,I_t^M\}$. The pairs $(I_{patch},I_t^i)$ and $(I_{patch},I_t^j)$ are fed into the style network as follows:
\begin{equation}
    I_{src}=\alpha_i\cdot I_{patch}+(1-\alpha_i)\cdot\mathrm{Net}_{s}(I_{patch},I_t^i),
    \label{eq:content_weight_i}
\end{equation}
\begin{equation}
    I_{tar}=\alpha_j\cdot I_{patch}+(1-\alpha_j)\cdot\mathrm{Net}_{s}(I_{patch},I_t^j),
    \label{eq:content_weight_j}
\end{equation}
where $\mathrm{Net}_s(\cdot)$ denotes the style transfer network; $\alpha_i$ and $\alpha_j$ are content weights uniformly distributed over the interval $[0,1]$. A larger value indicates greater similarity to $I_{patch}$. Since the style transfer network does not control the smoothness of textures, we apply image smoothing \cite{xu2011image} to $I_{src}$ and $I_{tar}$ by
\begin{equation}
    I_{src},I_{tar}=\mathrm{Smooth}(I_{src},\beta_i),\mathrm{Smooth}(I_{tar},\beta_j),
    \label{eq:smooth_weight}
\end{equation}
where $\beta_i$ and $\beta_j$ are smoothing weights uniformly distributed within $[0,\beta]$. A higher value results in greater smoothness. Subsequently, the ground-truth offsets $O_{gt}$ are generated to warp $I_{src}$
\begin{equation}
    I_{src}=\mathrm{Warp}(I_{src},O_{gt}),
\end{equation}
where $\mathrm{Warp}(\cdot)$ is the homography transformation function. Note that each element in $O_{gt}$ belongs to the integer set $\{-p, -p+1, ..., p\}$, with $p$ representing the maximum perturbation. Finally, center patches of size $S\times S$ are extracted from $I_{src}$ and $I_{tar}$ to create unaligned image pairs. Homography estimation models are trained on these unaligned image pairs along with the ground-truth offsets in a supervised manner. The content and template datasets consist solely of unimodal color images.

\subsection{Cross-Scale and Color-Invariant Network}
Existing homography models extract multiscale features to iteratively refine estimation results. However, these models focus solely on intra-scale information while neglecting cross-scale information, which limits accuracy. Additionally, they integrate color information from the image pairs into the extracted features, thus degrading generalization performance. To address these challenges, we propose the Cross-Scale and Color-Invariant Network (CCNet). CCNet effectively fuses features from different scales and decouples color information from the fused features. It utilizes the iterative strategy to reduce estimation errors.

% Existing homography models extract multiscale features to implement the iterative strategy to refine the estimation results. However, they only utilize intra-scale information and ignore cross-scale information, limiting the accuracy. Besides, the color information of the image pairs is also integrated into the extracted features, thus degrading the generalization performance. To address these problems, we design a Cross-scale and Color-invariant Network (CCNet). CCNet fuses the features of different scales and decouples color information from the fused features. It employs the iterative strategy to reduce estimation errors. 

\textbf{Overall Structure.} {\cref{fig:overall_framework}(b) shows the structure of CCNet.} For $\forall i\in\{src,tar\}$, the multiscale features of $I_i$ are extracted by
\begin{equation}
    F_{i}^1, F_{i}^2, F_{i}^3=\mathrm{Extractor}(I_{i}),
\end{equation}
where $F_{i}^1$, $F_{i}^2$, and $F_{i}^3$ have spatial sizes of $S\times S$, $S/2\times S/2$, $S/4\times S/4$, respectively; $\mathrm{Extractor}(\cdot)$ is a multiscale feature extraction network fusing cross-scale information. These extracted features are then passed through convolutional layers designed to eliminate color information. The resulting features are color-invariant and fed into the iterative homography estimation module for predicting offsets.

\textbf{Multiscale Feature Extractor.} \cref{fig:overall_framework}(e) illustrates the architecture of the extractor. For $\forall i\in\{src,tar\}$, the input image $I_i$ is processed through a convolutional layer and a residual block to extract the shallow feature $F^1\in\mathbb{R}^{C\times S\times S}$. Then, deeper features with reduced resolution are extracted and aggregated with the shallow feature
\begin{equation}
    F_i^2=\mathrm{ResBlock}(\mathrm{ResBlock}_{\downarrow}(F^1_i)\circ\mathrm{MaxPool}_{\downarrow}(F^1_i)),
\end{equation}
where $\mathrm{ResBlock}(\cdot)$ and $\mathrm{MaxPool}(\cdot)$ denote the residual block and max pooling, respectively; $``\downarrow"$ means that the resolution is decreased by half; $``\circ"$ denotes concatenation along the channel dimension. The third feature $F^3_i$ is obtained using the same operations. In this way, the cross-scale information is integrated in a top-to-bottom fashion. Besides, the information is also integrated in a bottom-to-top direction. Specifically, the spatial size of $F^3_i$ is increased by  a factor of two and then fused with $F^2_i$ by
\begin{equation}
    F^2_i=\mathrm{Conv}(F^2_i\circ\mathrm{Up}(F^3_i)),
\end{equation}
where $\mathrm{Up}(\cdot)$ denotes the nearest interpolation operation. The cross-scale information is also integrated into $F^1_i$ using the same methods.

\textbf{Color Decoupling.} \cref{fig:overall_framework}(c) illustrates the flowchart for color decoupling. For $\forall i\in\{src,tar\}$ and $\forall j\in\{1,2,3\}$, the multiscale feature $F^j_i$ is processed through convolutional layers to construct the color representation $F_{color}^{j,i}$ and the color-invariant feature $F_{invar}^{j,i}$. Inspired by \cite{pathak2025colors}, we facilitate this process with two loss functions. The first loss function focuses on color reconstruction and is defined as
\begin{equation}
    L_{color}^{j,i}=\mathrm{MSE}(\mathrm{Net}_{c}(F_{color}^{j,i}), \mathrm{Hist}(I_i)),
    \label{eq:color_recons}
\end{equation}
where $\mathrm{Hist}(\cdot)$ is a color histogram function; $\mathrm{Net}_{c}(\cdot)$ is a network that reconstructs the color information from the latent feature; $\mathrm{MSE}(\cdot,\cdot)$ denotes the L2 loss. The second loss function addresses color decoupling and is computed as
\begin{equation}
    L_{dis}^{j,i}=\|\mathrm{CosSim}(F^{j,i}_{color},F^{j,i}_{invar})\|_1,
    \label{eq:color_dis}
\end{equation}
where $``\|\cdot\|_1"$ and $\mathrm{CosSim}(\cdot,\cdot)$ represent the L1 loss and cosine similarity function, respectively. Minimizing this loss function encourages the color-invariant feature $F^{j,i}_{invar}$ to be orthogonal to the color feature $F^{j,i}_{color}$. Their correlations are minimized to decouple the color information from $F^{j,i}_{invar}$.

%%%%%%%%%%%%%%%%%%%%%%%%%%%%%%%%%%%%%%%%% figure %%%%%%%%%%%%%%%%%%%%%%%%%%%%%%%%%%%%%%%%
\begin{figure*}[t]
\centering
    \includegraphics[width=0.95\linewidth]{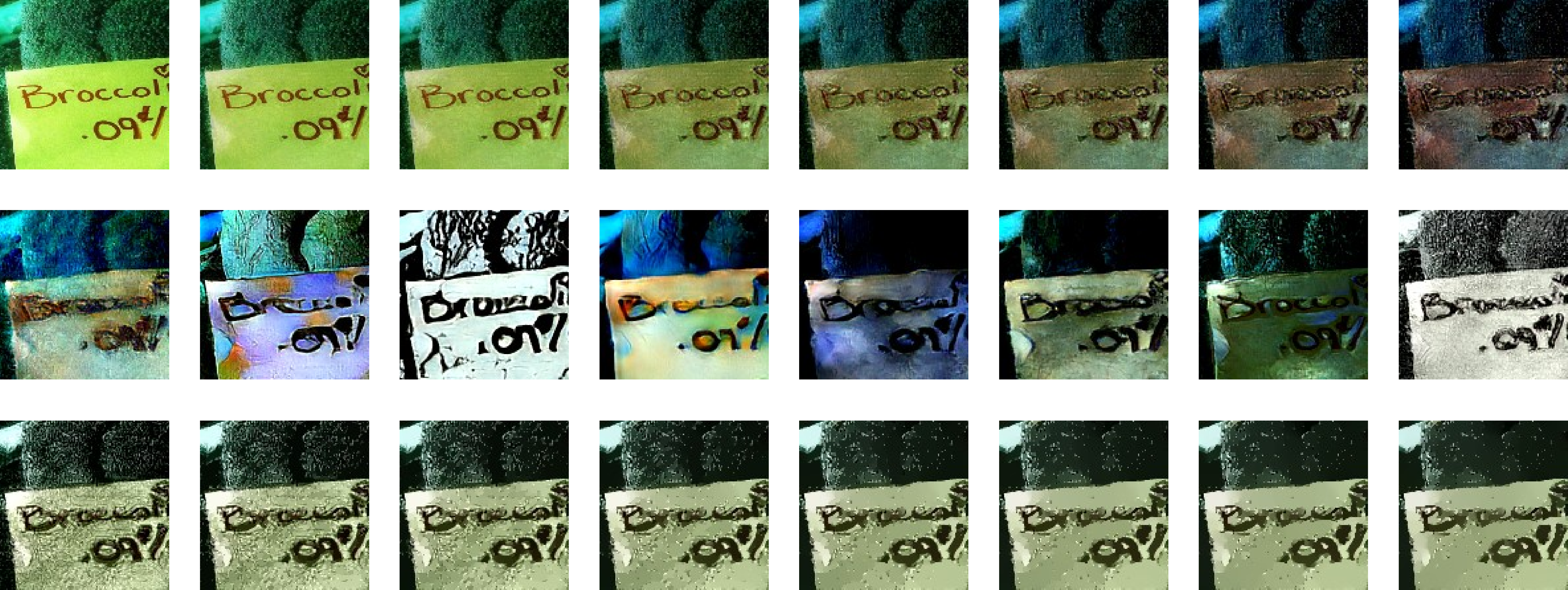}
    \caption{Examples of synthetic data. The first row presents the synthesis results with various content weights. The second row shows the results with different template images. The third row displays the examples with different smoothing weights. The weights become smaller and greater from left to right for the first and third rows, respectively.}
    \label{fig:example_synthesis}
\end{figure*}
%%%%%%%%%%%%%%%%%%%%%%%%%%%%%%%%%%%%%%%%%%%%%%%%%%%%%%%%%%%%%%%%%%%%%%%%%%%%%%%%%%%%%%%%

\textbf{Homography Estimation.} \cref{fig:overall_framework}(d) illustrates the iterative homography estimation module. It utilizes the color-invariant features to estimate the offsets at two levels. For $\forall j\in\{1,2,3\}$, the estimation result derived from $F^{src,j}_{invar}$ and $F^{tar,j}_{invar}$ can be expressed as
\begin{equation}
    O_{pred}^j=O_{pred}^{j,K}+O_{pred}^{j+1},
\end{equation}
where $O_{pred}^{j,K}$ is the residual offset output by the estimation block after $K$ iterations and $O_{pred}^4$ is initialized as a zero matrix. The first and second levels respectively utilize $O_{pred}^{j+1}$ and $O_{pred}^{j,k-1}$ to assist in estimating $O_{pred}^{j,k}$ for $\forall k\in\{1,2,...,K\}$. $O_{pred}^{j,0}$ is also set to a zero matrix. The color-invariant features $F^{src,j}_{invar}$ and $F^{tar,j}_{invar}$ are used throughout the iterations. For simplicity, we denote $F^j_{src}$ and $F^j_{tar}$ as $F^{src,j}_{invar}$ and $F^{tar,j}_{invar}$, respectively. To derive $O_{pred}^{j,k}$, $F^j_{src}$ is first warped by
\begin{equation}
    F^{j,k}_{src}=\mathrm{Warp}(F^{j}_{src},O_{pred}^{j,k-1}+O_{pred}^{j+1}),
\end{equation}
Then, the similarity between the target and warped source image features is calculated as follows:
\begin{equation}
    \begin{split}
        &C^{j,k}(u,v,m,n)\\
        &=\sum_{u=-r}^{r}\sum_{v=-r}^{r}{F^{j,k}_{src}}^T(m+u,n+v)\cdot F^{j}_{tar}(m,n),
    \end{split}
\end{equation}
where $(m,n)$ denotes the coordinates of a feature vector; $``T"$ represents the transpose operation; $u$ and $v$ enable the capture of contextual similarities; $r$ is the searching radius. The resulting 4D tensor $C^{j,k}$ encodes the contextual correlations between the target and source images. $C^{j,k}$ is reshaped into the size of $(2\cdot r+1)^2\times S/2^{j-1}\times S/2^{j-1}$ and input into the estimation block to update the residual offset
\begin{equation}
    O_{pred}^{j,k}=\mathrm{Net}_h(C^{j,k})+O_{pred}^{j,k-1},
\end{equation}
where $\mathrm{Net}_h(\cdot)$ comprises convolutional layers. $O_{pred}^1$ is the final output of the iterative homography estimation module.

\subsection{Loss Function}
Since the synthetic data includes the ground-truth offsets, the model is trained in a supervised manner. The loss function is defined as
\begin{equation}
    L=L_{pred}+\lambda\cdot\sum_{i\in\{src,tar\}}\sum_{j=1}^{3}\left( L_{color}^{j,i}+L_{dis}^{j,i}\right),
    \label{eq:loss}
\end{equation}
\begin{equation}
    L_{pred}=\sum_{O_{pred}\in\mathcal{O}}\|O_{pred}-O_{gt}\|_1,
\end{equation}
where $\lambda$ is a hyperparameter; $\mathcal{O}$ represents the set $\{O_{pred}^{j,k}+O_{pred}^{j+1}|j\in\{1,2,3\},k\in\{1,2,...,K\}\}$; The terms $L^{j,i}_{color}$ and $L^{j,i}_{dis}$ are defined in \cref{eq:color_recons,eq:color_dis}, respectively. The term $L_{pred}$ encourages the predicted offsets to approximate the ground truth as closely as possible. The other two terms in \cref{eq:loss} enforce the decoupling of color information from the features utilized for homography estimation.

\section{Experiments}
\label{sec:experiments}

\subsection{Experimental Settings}
\textbf{Datasets.} The proposed training data synthesis method and the homography estimation network were evaluated across four datasets: GoogleMap \cite{zhao2021deep}, GoogleEarth \cite{zhao2021deep}, RGB-NIR \cite{brown2011multi}, and PDSCOCO \cite{koguciuk2021perceptual}. The GoogleMap dataset includes paired images featuring both Google Maps and satellite map styles. The GoogleEarth dataset consists of images captured in Greater Boston throughout various seasons. The RGB-NIR dataset contains pairs of an RGB image and a near-infrared image. PDSCOCO features images that differ in brightness, saturation, contrast, and hue.

\textbf{Implementations.} Our training data synthesis method and homography estimation network are implemented using PyTorch. The training is conducted on a single RTX A6000 GPU. The AdamW is employed for parameter updates. For data synthesis, IEContraAST \cite{chen2021artistic} is employed as the style transfer network. Content images are sourced from the MSCOCO dataset \cite{lin2014microsoft} to enable zero-shot estimation. Template images are sampled from the Painter by Numbers dataset \cite{kaggle_painterbynumbers}. The training iterations, learning rate, $\beta$, $\lambda$, batch size, $K$, and $r$ are set to $1.2\times10^{5}$,  $4\times 10^{-4}$, $1\times 10^{-3}$, 0.5, 16, 2, and 4, respectively.

% \textbf{Baselines.} The baselines encompass supervised and unsupervised methods. The supervised methods include DHN \cite{detone2016deep}, MHN \cite{le2020deep}, IHN \cite{cao2022iterative}, RHWF \cite{cao2023recurrent}, and MCNet \cite{zhu2024mcnet}. The unsupervised methods consist of UDHN \cite{nguyen2018unsupervised}, CA-UDHN \cite{zhang2020content}, SCPNet \cite{zhang2024scpnet}, AltO \cite{song2024unsupervised}, and SSHNet \cite{yu2025sshnet}.
\textbf{Baselines.} The baselines encompass supervised and unsupervised methods. The supervised methods include DHN \cite{detone2016deep}, MHN \cite{le2020deep}, IHN \cite{cao2022iterative}, and MCNet \cite{zhu2024mcnet}. The unsupervised methods consist of UDHN \cite{nguyen2018unsupervised}, CA-UDHN \cite{zhang2020content}, SCPNet \cite{zhang2024scpnet}, AltO \cite{song2024unsupervised}, and SSHNet \cite{yu2025sshnet}.

\textbf{Metrics.} Following prior work, the mean average corner error (MACE) is adopted to evaluate the accuracy of homography estimation. A smaller MACE value indicates higher estimation accuracy.

%%%%%%%%%%%%%%%%%%%%%%%%%%%%%%%%%%%%%%%% table %%%%%%%%%%%%%%%%%%%%%%%%%%%%%%%%%%%%%%
\begin{table}[t]
\caption{Cross-dataset evaluation for GoogleMap. $``*"$ means that the model is trained on our synthetic data. $``+"$ indicates that the data synthesis method is applied to GoogleMap for training.}
\label{tab:cross_ggmap}
\begin{tabular}{cccc}
\hline
Method & GoogleEarth & RGB-NIR & PDSCOCO  \\ \hline

\rowcolor{myrowcolor}
DHN \cite{detone2016deep}    & 27.653      & 26.605  &29.856 \\

$\textrm{DHN}^*$   &22.933       &13.864   &14.173  \\

\rowcolor{myrowcolor}
MHN \cite{le2020deep}    & 39.474      & 30.372  & 33.490\\

$\textrm{MHN}^*$   & 3.110       & 7.549  &4.251   \\

\rowcolor{myrowcolor}
IHN \cite{cao2022iterative}    & 3.038       & 12.491  & 5.352\\

\rowcolor{myrowcolor2}
$\textrm{IHN}^+$     & 2.770       & 7.456  & 4.481  \\

$\textrm{IHN}^*$     & 1.853       & 5.647  &1.684   \\

% \rowcolor{myrowcolor}
% RHWF \cite{cao2023recurrent}   & 2.452       & 12.134  & 5.056\\

% $\textrm{RHWF}^*$      & 1.678       & 4.401  &1.455   \\

\rowcolor{myrowcolor}
MCNet \cite{zhu2024mcnet}   & 20.518      & 16.557  & 8.202\\ 

\rowcolor{myrowcolor2}
$\textrm{MCNet}^+$ & 4.198        &10.450   &8.489\\

$\textrm{MCNet}^*$&1.402        &5.239   &1.423\\

\hline

SCPNet \cite{zhang2024scpnet} & 5.508       & 10.918  &4.862 \\
AltO \cite{song2024unsupervised}   & 9.089       & 18.651  &12.398 \\
SSHNet \cite{yu2025sshnet} & 24.952      & 30.535  &25.486 \\
\hline
\end{tabular}
\end{table}
%%%%%%%%%%%%%%%%%%%%%%%%%%%%%%%%%%%%%%%%%%%%%%%%%%%%%%%%%%%%%%%%%%%%%%%%%%%%%%%%%%%%%%%%

%%%%%%%%%%%%%%%%%%%%%%%%%%%%%%%%%%%%%%%% table %%%%%%%%%%%%%%%%%%%%%%%%%%%%%%%%%%%%%%
% \usepackage{multirow}
\begin{table}[t]
\caption{Cross-dataset evaluation for GoogleEarth. $``*"$ means that the model is trained on our synthetic data. $``+"$ indicates that the data synthesis method is applied to GoogleEarth for training.}
\label{tab:cross_ggearth}
\begin{tabular}{cccc}
\hline
Method & GoogleMap & RGB-NIR & PDSCOCO  \\ \hline

\rowcolor{myrowcolor}
  DHN \cite{detone2016deep}    & 24.604    & 21.937  &21.348 \\
  
$\textrm{DHN}^*$     &11.321       &13.864   &14.173   \\

\rowcolor{myrowcolor}
  MHN \cite{le2020deep}    & 26.806    & 20.581  &11.518 \\
  
$\textrm{MHN}^*$    & 11.218       & 7.549  &4.251   \\

\rowcolor{myrowcolor}
  IHN \cite{cao2022iterative}    & 14.962    & 12.616  &5.278 \\
  
\rowcolor{myrowcolor2}
$\textrm{IHN}^+$    & 7.223       & 5.647  &1.684   \\
  
$\textrm{IHN}^*$    & 5.303       & 8.336  &2.449   \\

% \rowcolor{myrowcolor}
%   RHWF \cite{cao2023recurrent}   & 14.338    & 11.503  &4.685 \\
  
% $\textrm{RHWF}^*$   & 4.212         & 4.401  &1.455   \\

\rowcolor{myrowcolor}
  MCNet \cite{zhu2024mcnet}  & 10.799    & 9.982   &3.753 \\ 
  
\rowcolor{myrowcolor2}
$\textrm{MCNet}^+$  &6.334        &7.048   &1.607   \\
  
$\textrm{MCNet}^*$  &5.093        &5.239   &1.423   \\

\hline

  SCPNet \cite{zhang2024scpnet} & 28.291    & 14.281  & 4.460\\
  AltO \cite{song2024unsupervised}   & 17.835    & 17.306  & 9.566\\
  SSHNet \cite{yu2025sshnet} & 21.335    & 17.419  &10.114 \\ 
\hline
\end{tabular}
\end{table}
%%%%%%%%%%%%%%%%%%%%%%%%%%%%%%%%%%%%%%%%%%%%%%%%%%%%%%%%%%%%%%%%%%%%%%%%%%%%%%%%%%%%%%%%

%%%%%%%%%%%%%%%%%%%%%%%%%%%%%%%%%%%%%%%%%%% figure %%%%%%%%%%%%%%%%%%%%%%%%%%%%%%%%%%%%%%%
% \usepackage{multirow}
\begin{table}[t]
\caption{Cross-dataset evaluation for RGB-NIR. $``*"$ means that the model is trained on our synthetic data. $``+"$ indicates that the data synthesis method is applied to RGB-NIR for training.}
\label{tab:cross_rgbnir}
\begin{tabular}{cccc}
\hline
Method & GoogleMap & GoogleEarth & PDSCOCO \\ \hline

\rowcolor{myrowcolor} 
DHN \cite{detone2016deep}    & 15.188    & 18.288   &19.962    \\

$\textrm{DHN}^*$     &11.321     &22.933   &14.173   \\

\rowcolor{myrowcolor}
MHN \cite{le2020deep}    & 16.931    & 4.265    &4.927    \\

$\textrm{MHN}^*$    & 11.218    & 3.110      &4.251   \\

\rowcolor{myrowcolor}
IHN \cite{cao2022iterative}    & 10.629    & 2.095    &2.873    \\

\rowcolor{myrowcolor2}
$\textrm{IHN}^+$     & 3.768    & 1.894    &1.901   \\

$\textrm{IHN}^*$     & 5.303    & 1.853    &1.684   \\

% \rowcolor{myrowcolor}
% RHWF \cite{cao2023recurrent}   & 8.647    & 1.822     &2.388  \\

% $\textrm{RHWF}^*$   & 4.212     & 1.678    &1.455   \\

\rowcolor{myrowcolor}
MCNet \cite{zhu2024mcnet}  & 6.312     & 1.843    & 1.451   \\ 

\rowcolor{myrowcolor2}
$\textrm{MCNet}^+$  &3.605      &1.569        &2.245   \\

$\textrm{MCNet}^*$  &5.093      &1.402        &1.423   \\
\hline

SCPNet \cite{zhang2024scpnet} & 19.109    & 10.618   &3.749    \\
AltO \cite{song2024unsupervised}   & 15.756    & 5.181    &7.760    \\
SSHNet \cite{yu2025sshnet} & 23.680    & 2.848    &3.879    \\
\hline
               
\end{tabular}
\end{table}
%%%%%%%%%%%%%%%%%%%%%%%%%%%%%%%%%%%%%%%%%%%%%%%%%%%%%%%%%%%%%%%%%%%%%%%%%%%%%%%%%%%%%%%%

%%%%%%%%%%%%%%%%%%%%%%%%%%%%%%%%%%%%%% figure %%%%%%%%%%%%%%%%%%%%%%%%%%%%%%%%%%%%%%%%%%%
\begin{table}[t]
\caption{Cross-dataset evaluation for PDSCOCO. $``*"$ means that the model is trained on our synthetic data. $``+"$ indicates that the data synthesis method is applied to PDSCOCO for training.}
\label{tab:cross_pdscoco}
\begin{tabular}{cccc}
\hline
Method & GoogleMap & GoogleEarth & RGB-NIR \\ \hline

\rowcolor{myrowcolor}
DHN \cite{detone2016deep}    &18.315    & 11.961   &12.669    \\

$\textrm{DHN}^*$     &11.321     &22.933       &13.864    \\

\rowcolor{myrowcolor}
MHN \cite{le2020deep}    &22.822    &5.694    &11.272    \\

$\textrm{MHN}^*$    & 11.218    & 3.110       & 7.549  \\

\rowcolor{myrowcolor}
IHN \cite{cao2022iterative}    &21.727    & 2.156    &10.745    \\

% \rowcolor{myrowcolor2}
% $\textrm{IHN}^+$    & 4.989    &1.824       & 5.569  \\

$\textrm{IHN}^*$    & 5.303    &1.853       & 5.647  \\

% \rowcolor{myrowcolor}
% RHWF \cite{cao2023recurrent}   &25.219    &2.711     &11.451  \\

% $\textrm{RHWF}^*$   & 4.212     & 1.678       & 4.401  \\

\rowcolor{myrowcolor}
MCNet \cite{zhu2024mcnet}  &22.276     & 2.083   &9.186    \\

% \rowcolor{myrowcolor2}
% $\textrm{MCNet}^+$  &5.095      &1.406        &4.641   \\

$\textrm{MCNet}^*$  &5.093      &1.402        &5.239    \\
\hline

SCPNet \cite{zhang2024scpnet} &6.553    & 11.673   &12.528    \\
AltO \cite{song2024unsupervised}   & 27.060    & 5.945   &14.739    \\
SSHNet \cite{yu2025sshnet} & 23.586    & 2.105    &11.010    \\
      
\hline
\end{tabular}
\end{table}
%%%%%%%%%%%%%%%%%%%%%%%%%%%%%%%%%%%%%%%%%%%%%%%%%%%%%%%%%%%%%%%%%%%%%%%%%%%%%%%%%%%%%%%%

\subsection{Training Data Synthesis}

\textbf{Visualization.} \cref{fig:example_synthesis} showcases the results of rendering the same image with various parameters. As the content weights decrease, the images exhibit greater divergence from the original in terms of texture and color. Applying different styles can result in a variety of textures and colors. Additionally, a larger smoothing weight yields smoother textures. Despite this diversity, the structural information of the original image is preserved. This combination of structural consistency and appearance diversity enhances the model's ability to generalize across various modalities.

\begin{table*}[t]
\centering
\caption{Within-dataset and zero-shot evaluation results (MACE $\downarrow$) of our training data synthesis method and the homography estimation network. The best result is written in bold while the second-best one is underlined. ``+" indicates that the models are trained on the dataset augmented by our synthesis method. ``Zero-shot'' means that the models are trained on our synthetic method.}
\label{tab:within_dataset}
\begin{tabular}{ccccccc}
\hline
Type         & Method  & GoogleMap & GoogleEarth & RGB-NIR &PDSCOCO \\
\hline
% Traditional  & SIFT    &           &             &         \\
%              & ORB     &           &             &         \\
%              & RIFT    &           &             &         \\

\multirow{7}{*}{Supervised}   
& DHN \cite{detone2016deep}     & 3.556     & 7.735       & 12.855 &  11.937 \\
& MHN \cite{le2020deep}     & 1.561     & 2.061       & 7.078  & 5.142 \\
& IHN \cite{cao2022iterative}    & 1.025     & 1.147       & 4.588  & 1.445  \\
% & RHWF \cite{cao2023recurrent}   & 0.755     & 1.029       & 3.820   &  1.647 \\
& MCNet \cite{zhu2024mcnet}  & \underline{0.261}     & \underline{0.577}       & \underline{3.226}  & \underline{1.062}  \\
& CCNet (Ours)    & \textbf{0.184}     & \textbf{0.526}       & \textbf{2.992} &\textbf{1.001}  \\

\rowcolor{myrowcolor2}
& $\textrm{IHN}^{+}$    & 1.515     & 1.236      & 3.552  & 1.498\\
\rowcolor{myrowcolor2}
& $\textrm{MCNet}^{+}$    & 0.992     & 0.776      & 3.146  & 1.348 \\
\hline

\multirow{5}{*}{Unupervised} 
& UDHN \cite{nguyen2018unsupervised}   & 24.713    & 21.248      & 25.452 & 25.684 \\
& CA-UDHN \cite{zhang2020content} & 24.557    & 23.761      & 24.297 & 24.875  \\
& SCPNet \cite{zhang2024scpnet} & 4.364     & 2.794       & 10.618  & 9.448  \\
& AltO \cite{song2024unsupervised}   & 4.739     & 3.174       & 7.897 &  4.979   \\
& SSHNet \cite{yu2025sshnet} & 1.394     & 5.888       & 6.743  & 1.610  \\
\hline

\multirow{5}{*}{Zero-shot}    
& DHN \cite{detone2016deep}     &11.321     &22.933       &13.864   &14.173  \\
& MHN \cite{le2020deep}    & 11.218    & 3.110       & 7.549  &4.251 \\
& IHN \cite{cao2022iterative}    & 5.303    & 1.853       & 5.647  &1.684  \\
% & RHWF \cite{cao2023recurrent}   & 4.212     & 1.678       & 4.401  &1.455   \\
& MCNet \cite{zhu2024mcnet}  &5.093      &1.402        &5.239   &1.423 \\
& CCNet (Ours)    &\textbf{4.383}      &\textbf{1.399}   &\textbf{4.461}  & \textbf{1.368} 

\\
\hline 
\end{tabular}
\end{table*}
%%%%%%%%%%%%%%%%%%%%%%%%%%%%%%%%%%%%%%%%%%%%%%%%%%%%%%%%%%%%%%%%%%%%%%%%%%%%%%%%%%%%%%%%

\textbf{Cross-Dataset Evaluation.} We first assess the effectiveness of the proposed data synthesis method through cross-dataset evaluation. Higher accuracy signifies better generalization performance.  \cref{tab:cross_ggmap,tab:cross_ggearth,tab:cross_rgbnir,tab:cross_pdscoco} present the results for the baselines trained on GoogleMap, GoogleEarth, RGB-NIR, and PDSCOCO. These tables also include results for training on our synthetic data, which enables zero-shot estimation. The findings indicate that the baselines exhibit unsatisfactory generalization performance when trained on existing datasets, particularly with GoogleMap, GoogleEarth, and PDSCOCO. In contrast, the RGB-NIR dataset affords improved generalization. The near-infrared wavelength captures less texture and color information, allowing the structural information to dominate the images. Training on our synthetic data enhances the generalization performance of the baselines. The generalization performance of zero-shot DHN declines in three cases due to its weaker learning capability compared to other methods. Excluding these cases, improvements facilitated by our data synthesis method range from 1.93\% to 93.17\%, with over 50\% improvement observed in roughly half of the cases. Besides, PDSCOCO is also derived from MSCOCO and exhibits variations in brightness, saturation, hue, and contrast. The generalization performance remains unsatisfactory when training models on it. This suggests that our synthesis strategy contributes to better generalization beyond the variability offered by the content images in MSCOCO.

\textbf{Augmentation.} We regard the synthesis method as a form of augmentation and apply it to existing datasets for further evaluation. IHN and MCNet are selected for testing due to their superior estimation capabilities. The results are presented in \cref{tab:cross_ggmap,tab:cross_ggearth,tab:cross_rgbnir}. Although estimation accuracy decreases in two cases, the synthesis method generally enhances the generalization performance by 8.82\% to 79.54\%. This indicates that our synthesis strategy is beneficial for generalization, irrespective of the content images used.

\textbf{Within-Dataset Evaluation.} \cref{tab:within_dataset} reports the within-dataset and zero-shot accuracy of the baselines. Both supervised and unsupervised methods are trained and tested on the same dataset, while the zero-shot methods are trained on data synthesized from MSCOCO. As observed, the supervised methods generally achieve the highest within-dataset performance, as they leverage ground-truth offsets to adapt to the specific modalities of the training dataset. Applying our synthesis method to existing datasets enhances generalization but also results in a decrease in accuracy. IHN and MCNet are dissuaded by the synthetic data from fully utilizing modality information to infer the offsets. Furthermore, our synthetic data enhances zero-shot estimation due to its diversity in textures and colors. Conversely, the lack of modality information results in lower accuracy for within-dataset evaluations compared to supervised methods. Notably, the zero-shot baselines exhibit estimation performance comparable to that of unsupervised methods in within-dataset evaluations while demonstrating superior generalization capabilities. The unsupervised method SSHNet achieves significantly higher accuracy on GoogleMap than the zero-shot methods, as it employs image translation techniques to unify the modalities of image pairs. However, as shown in \cref{tab:cross_ggmap}, this approach also substantially degrades generalization performance.

{The supplementary material reports more experimental results for the training data synthesis method.}

%%%%%%%%%%%%%%%%%%%%%%%%%%%%%%%%%%%%%%% figure %%%%%%%%%%%%%%%%%%%%%%%%%%%%%%%%%%%%%%%%%
\begin{figure*}[t]
    \centering
    \includegraphics[width=0.97\linewidth]{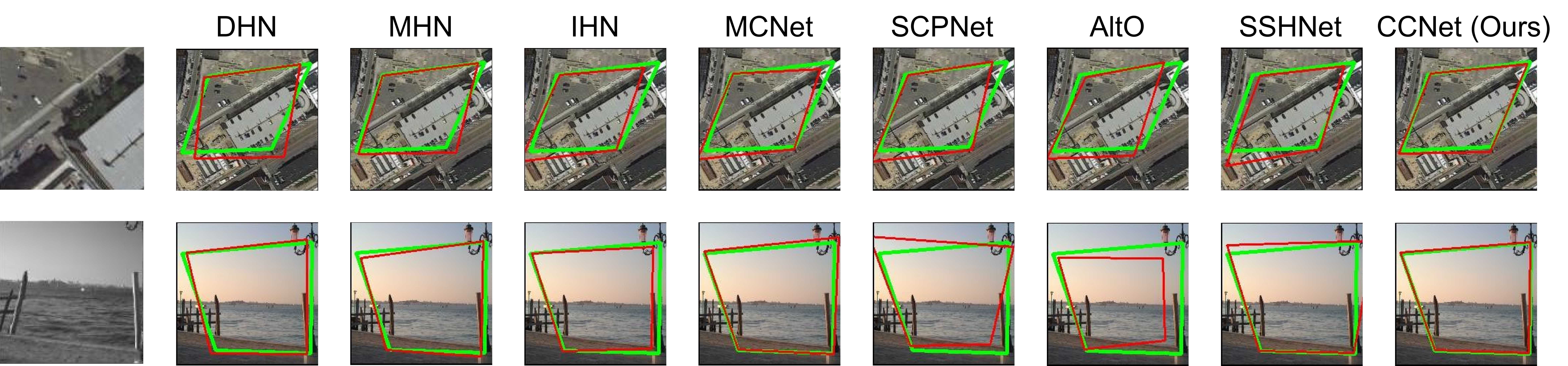}
    \caption{Visualization results of within-dataset evaluation. The first row presents the estimation results from GoogleEarth, while the second row features results from RGB-NIR. The first column displays the source image to be warped, while the other columns demonstrate the target images. Greater similarity between the red and green quadrilaterals indicates higher accuracy in the estimation.}
    \label{fig:visual_within_dataset}
\end{figure*}
%%%%%%%%%%%%%%%%%%%%%%%%%%%%%%%%%%%%%%%%%%%%%%%%%%%%%%%%%%%%%%%%%%%%%%%%%%%%%%%%%%%%%%%%

%%%%%%%%%%%%%%%%%%%%%%%%%%%%%%%%%%%%%%% figure %%%%%%%%%%%%%%%%%%%%%%%%%%%%%%%%%%%%%%%%%
\begin{figure}[t]
    \centering
    \includegraphics[width=1\linewidth]{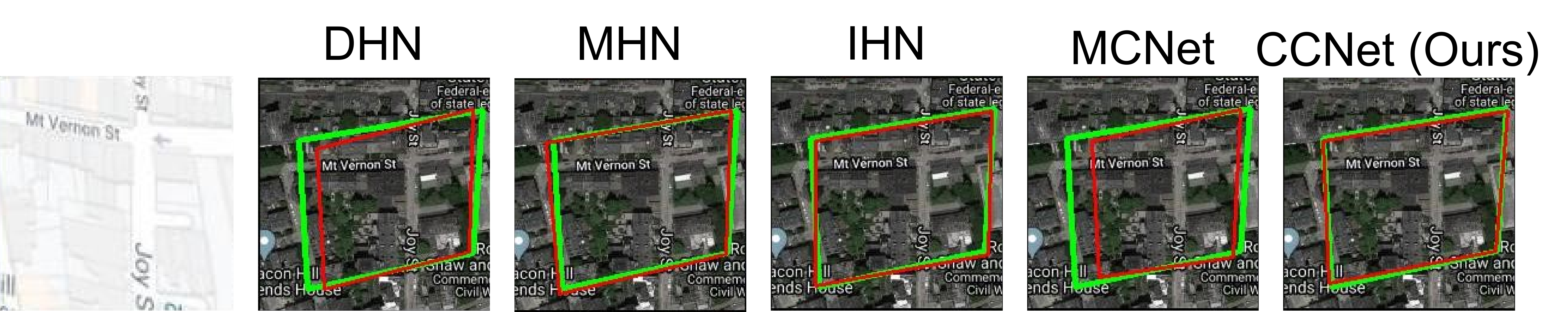}
    \caption{Visualization results of zero-shot evaluation using testing images from GoogleMap. The first column displays the source image to be warped. A greater similarity between the red and green quadrilaterals indicates higher accuracy in estimation.}
    \label{fig:visual_zero_shot}
\end{figure}
%%%%%%%%%%%%%%%%%%%%%%%%%%%%%%%%%%%%%%%%%%%%%%%%%%%%%%%%%%%%%%%%%%%%%%%%%%%%%%%%%%%%%%%%

\subsection{Cross-Scale and Color-Invariant Network}
\textbf{Quantitative Results.} We compare our CCNet with the baselines when they are trained on the four datasets and our synthetic data. \cref{tab:within_dataset} shows the comparison results. As can be observed, our CCNet outperforms the baselines in both within-dataset and zero-shot estimation tasks. In the within-dataset evaluation, CCNet achieves higher accuracy than the second-best method, with improvements of 29.50\%, 8.83\%, 7.25\%, and 5.74\% on GoogleMap, GoogleEarth, RGB-NIR, and PDSCOCO, respectively. These enhancements underscore CCNet's superior capability in multimodal homography estimation. For zero-shot evaluation, CCNet shows improvements of 13.94\%, 0.21\%, 14.85\%, and 3.87\% on the four datasets. CCNet effectively integrates the cross-scale information into the features extracted for homography estimation, leading to enhanced results. Additionally, by decoupling color from the features, CCNet minimizes the negative effects associated with color information when processing multimodal image pairs.

\textbf{Qualitative Results.} \cref{fig:visual_within_dataset} presents the visual results of within-dataset evaluation on GoogleEarth and RGB-NIR, while \cref{fig:visual_zero_shot} showcases the visualizations for zero-shot evaluation on GoogleMap. As demonstrated, our CCFNet estimates the offsets more accurately than the baselines across both within-dataset and zero-shot evaluations.

\textbf{Computational Costs.} \cref{tab:computational_costs} presents the runtime and model size of various supervised methods. As indicated in \cref{tab:within_dataset} and \cref{tab:computational_costs}, our CCFNet demonstrates superior homography estimation and generalization capabilities, requiring only a slight increase in runtime and parameters.

{The supplementary material provides additional experimental results for CCNet.}

%%%%%%%%%%%%%%%%%%%%%%%%%%%%%%%%%%%%%% figure %%%%%%%%%%%%%%%%%%%%%%%%%%%%%%%%%%%%%%%%%%%
\begin{table}[t]
\caption{Computational costs of supervised methods and our network. The first row displays the runtime (in milliseconds) while the second row presents the model size (in megabytes).}
\label{tab:computational_costs}
\centering
\resizebox{1\linewidth}{!}{
\begin{tabular}{ccccc}
\hline
DHN \cite{detone2016deep}   & MHN \cite{le2020deep} & IHN  \cite{cao2022iterative} &MCNet \cite{zhu2024mcnet} &CCNet \\ \hline
11.21 &10.92 &26.14 &31.01 &32.73\\
34.20 &2.57 &0.76 &0.85 &1.21\\

\hline
\end{tabular}
}
\end{table}
\section{Conclusion}
In this paper, we introduce a training data synthesis method aimed at enhancing generalization in multimodal homography estimation. This method produces various unaligned image pairs by rendering the same image with different textures and colors, along with their corresponding ground-truth offsets. While these images exhibit diverse appearances, they retain the structural information of the original image. Therefore, a homography estimation model can be trained on these data in a supervised manner to achieve zero-shot estimation. Additionally, the synthesis strategy can be applied to existing datasets to improve generalization, albeit with a slight compromise in within-dataset performance. We also design a network to enhance homography estimation accuracy. This network fully leverages cross-scale information to enhance the estimation results. Color information is decoupled from the features to boost the network's ability in multimodal image processing. Extensive experiments are conducted to demonstrate the effectiveness of the proposed data synthesis method and the superiority of our homography estimation network.

\section*{Acknowledgements}
This work was funded in part by the Science and Technology Development Fund, Macau SAR (File no. 0050/2024/AGJ), by the University of Macau and University of Macau Development Foundation (File no. MYRG-GRG2024-00181-FST-UMDF).
{
    \small
    \bibliographystyle{ieeenat_fullname}
    \bibliography{main}
}



\end{document}